\documentclass{article}

\usepackage[english]{babel}

\usepackage[letterpaper,top=2cm,bottom=2cm,left=3cm,right=3cm,marginparwidth=1.75cm]{geometry}

\usepackage{amsmath,amssymb}
\usepackage[linesnumbered,ruled,vlined]{algorithm2e}
\usepackage{graphicx}
\usepackage[colorlinks=true, allcolors=blue]{hyperref}
\usepackage{authblk}
\usepackage[skip=10pt plus1pt]{parskip}
\usepackage{caption} 
\usepackage{breakcites}
\usepackage{adjustbox}

\title{Synthetic data generation method for data-free knowledge distillation in regression neural networks}
\author[1]{Tianxun Zhou}
\author[1]{Keng-Hwee Chiam}
\affil[1]{Bioinformatics Institute, Singapore}
\date{}

\begin{document}
\setlength{\parindent}{0pt}
\maketitle

\begin{abstract}
Knowledge distillation is the technique of compressing a larger neural network, known as the
teacher, into a smaller neural network, known as the student, while still trying to maintain the performance of the larger neural network as much as possible. Existing methods of knowledge distillation are mostly applicable for classification tasks. Many of them also require access to the data used to train the teacher model. To address the problem of knowledge distillation for regression tasks in the absence of original training data, the existing method uses a generator model trained adversarially against the student model to generate synthetic data to train the student model. In this study, we propose a new synthetic data generation strategy that directly optimizes for a large but bounded difference between the student and teacher model. Our results on benchmark experiments demonstrate that the proposed strategy allows the student model to learn better and emulate the performance of the teacher model more closely.
\end{abstract}

\section{Introduction}

In the recent decade, advances in algorithms, computational hardware and data availability have enabled significant developments in artificial neural networks and deep learning \cite{Lecun2015}. Neural networks models are now state-of-the-art in many fields of application including computer vision (\cite{OMahony2020, Chai2021}, natural language processing \cite{Otter2021}, and signal processing \cite{Purwins2019, Rim2020}. However, as models become increasingly larger in size measured by number of parameters, they too become computationally expensive to store and perform inference on. Large neural networks can be unusable for real world deployment scenarios where hardware may be limited, such as on mobile devices or microcontrollers, or when deployed for as a service to a large number of users such as web applications \cite{Cheng2018, Deng2020, Liu2022}.
\par
Knowledge distillation is a class of method to address this problem by distilling the predictive capabilities of a larger neural network into a smaller neural network, allowing for faster inference and lower memory requirements \cite{Gou2021}. There have been several knowledge distillation methods proposed in the past, typically requiring the original data that was used to train the teacher model. However, in many real-world applications, the original data may not be available for performing knowledge distillation to student models due to reasons such as data size and data privacy \cite{Chen2019, Gou2021}.
\par
To deal with such situations, data-free knowledge distillation methods have been proposed to allow distillation of knowledge without the original training data \cite{Hu2020, Lopes2017, Micaelli2019, Ye2020, Yoo2019}. Data-free knowledge distillation works by generating synthetic data and training the student model with these data and their teacher model predicted labels.
\par
Much of the existing research for knowledge distillation has been focused on classification tasks. However, regression tasks are common in many engineering applications \cite{Guo2021, Schweidtmann2021, Tapeh2022, Thai2022} and there are limited methods available on knowledge distillation for regression neural networks. Knowledge distillation for real-world regression applications include lightweight object detectors for remote sensing \cite{Yang2022}, turbine-scale wind power prediction \cite{Chen2022}. Recently, \cite{Kang2021} proposed the first data-free knowledge distillation method for regression where a generator model was trained in an adversarial manner to generate synthetic data. Motivated by the need for data-free model distillation on regression models in real world applications, in this work we investigate the behaviors of several synthetic data generation methods including random sampling and adversarial generator. 

The main contribution of this paper is to propose an improved method to generate synthetic data for data-free knowledge distillation of regression neural networks. This is achieved by
optimizing for a loss function defined using the student and teacher model predictions directly rather than implicitly through an additional generator model. Compared to existing methods, our synthetic data generation method can provide large difference in prediction between the student and teacher model while mimicking real data better. We demonstrate that this method for synthetic data generation can provide better performance than existing methods through experiments in 7 standard regression datasets, as well as on the MNIST handwritten digit dataset adapted for regression, and a real-world bioinformatics case study of protein solubility prediction. This study contributes in furthering the understanding of data-free distillation for regression and has potential applications in the deployment of distilled models
for many practical regression tasks in various engineering applications.
\par
\section{Related work}

\subsection{Knowledge distillation}

As neural networks become increasingly large in number of parameters, the deployment of such models faces a difficult challenge for applications such as mobile devices and embedded systems due to limitations in computational resources and memory \cite{Cheng2018, Deng2020}.  To address such problems, model compression through knowledge distillation has become an active area of research in recent years \cite{Liu2022, CHWang2022}. Knowledge distillation is the technique where knowledge learned by a larger teacher model is transferred to a smaller student model \cite{Gou2021, Hinton2015, Wang2022}. The main idea is that the student model mimics the teacher model to achieve a similar or even a superior performance.
\par
Various methods of knowledge distillation define and focus on different forms of knowledge. Following the nomenclature in \cite{Gou2021}, these can be largely grouped as response-based knowledge, feature-based knowledge, and relation-based knowledge. For response-based knowledge, outputs of the teacher model are used to supervise the training of the student model. For example, \cite{Hinton2015} uses soft targets from the logits output of the teacher model to train the student. For feature-based knowledge, outputs of intermediate layers, or feature maps learned by the teacher model can be to supervise the training of the student model. For example, \cite{Romero2014} trains the student model to match the feature activations of the teacher model. For relationship-based knowledge, the relationships between different layers or data samples are used. For example, \cite{Yim2017} uses the inner products between features from two layers to represent the relationship between different layers, while \cite{Chen2021} trains the student model to learn to preserve the similarity of samples’ feature embeddings in the intermediate layers of the teacher models.
\par
\subsection{Data-free knowledge distillation}

In some situations, access to the original data used to train the teacher model is not available due to issues such as privacy and legal reasons. Data-free knowledge distillation methods have been proposed to allow model distillation in the absence of original training data \cite{Chawla2021, Chen2019, Hu2020, Lopes2017, Micaelli2019, Nayak2021, Ye2020, Yoo2019}. This is achieved by generating synthetic data for training. Many methods achieve this by using generative adversarial networks (GAN) \cite{Chen2019, Hu2020, Micaelli2019, Ye2020, Yoo2019}. For example, \cite{Micaelli2019} train a generator model to generate synthetic images that maximizes the difference in prediction (measured by KL divergence) between the teacher and student models. The student model is then trained to minimize the difference on these synthetic images. When
applied to image classification datasets (SVHN and CIFAR-10 datasets), they were able to train student models with performance close to distillation using actual full training set.
\par
Other methods such as \cite{Lopes2017} make use of metadata collected during training of the teacher model, in the form of the layer activation records of the teacher model to reconstruct dataset for training the student model. 
\par
\subsection{Knowledge distillation for regression}

Most of the methods currently existing in knowledge distillation literature deal with classification problems. These methods generally are not immediately applicable to regression problems where the predictions are unbounded real values. For regression problems, \cite{Chen2017} uses a teacher bounded regression loss where the teacher’s predictions serve as an upper bound for the student model instead of using it directly as a target. This method was the first successful application of knowledge distillation to object detection problem and achieved strong results.
\par
\cite{Saputra2019} uses the teacher loss as a confidence score that assigns relative importance on the teacher prediction. This confidence score is then used in the main training task via an attentive imitation loss which adaptively down-weight the loss between the teacher and student model predictions when a particular teacher model’s prediction is not reliable. This method successfully performed knowledge distillation from a deep pose regression network, achieving a student model prediction close to the teacher model while reducing number of parameters by more than 90\%.
\par
\cite{Takamoto2020} uses a teacher outlier rejection loss, that rejects outliers in training samples based on the teacher model predictions. This method was applied to predict gaze angles from images of human face and was demonstrated to be effective even with noisy labels.
\par
\cite{Xu2022} proposed the Contrastive Adversarial Knowledge Distillation (CAKD) approach for time-series regression tasks. CAKD uses adversarial adaptation to align feature distributions between student and teacher models. To achieve better alignment on fine-grained features, contrastive learning is also used to increase similarity between the features extracted by teacher and student models for the same sample. The teacher model prediction is used as soft labels to guide the training for the student.
\par
\cite{Kang2021} introduced the first work, and to our knowledge the only work that addresses data-free knowledge distillation for regression, by using a generator model that generates synthetic datapoints that is trained adversarially together with the student model. The method was able to substantially outperform the baseline methods of random sampling and
data impression on various benchmark regression datasets. However, it was found that both the baseline methods and the generator method proposed here did not work well for deeper neural networks.
\par
\section{Material and methods}

\subsection{Overview of methods}

\begin{figure}
\centering
\includegraphics[width=\textwidth]{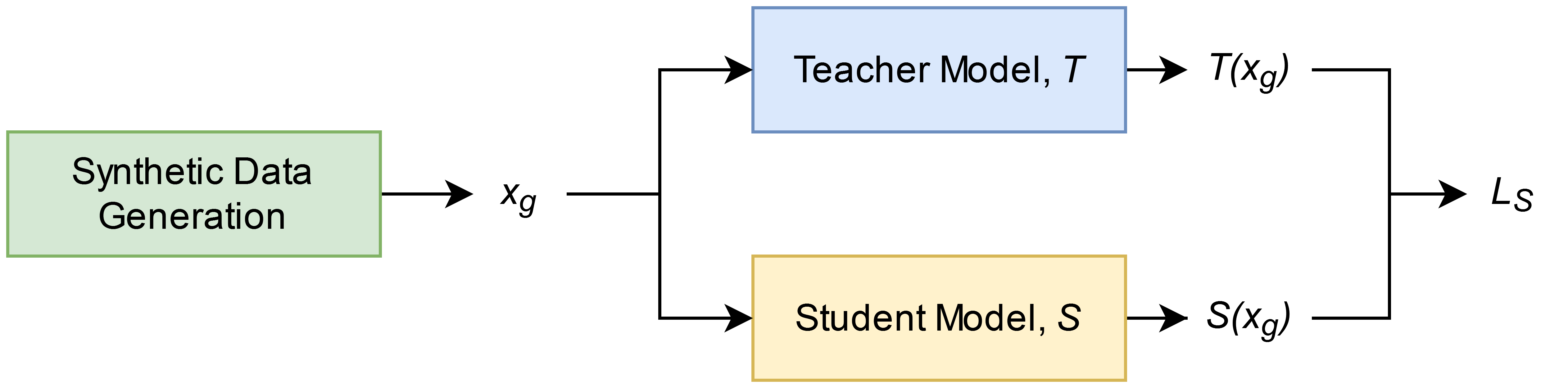}
\caption{\label{fig:fig1} Generic data-free knowledge distillation method}
\end{figure}

Given a trained teacher model T, and a student model $S_{\theta}$ parameterized by $\theta$, we generate synthetic data $x$ via some data generation method. The student model is trained by minimizing the student loss $L_S(x)$ defined in equation \ref{eq:1} using gradient descent. This generic method is illustrated in Figure \ref{fig:fig1}.

\begin{equation} \label{eq:1}
L_S(x) = (T(x) - S_{\theta}(x))^2
\end{equation}

The performance of the student model in mimicking the performance of the teacher is dependent on the representation strength $\theta$ of the student model, and the data $x$ used to train it, and the optimization process of minimizing student loss. Hence for a fixed student model architecture and training process, the synthetic data generation process plays the key role in determining the performance of the student model.

\subsection{Synthetic data generation methods}
Three types of synthetic data generation methods are investigated in this study: random sampling, generative model, and direct optimization.
\par
\subsubsection{Random sampling}
Synthetic data are generated by sampling randomly from an input distribution. To obtain the best results, the synthetic data should be drawn from a similar distribution as the actual data.
\par
In the case where actual input data have been standardized, random samples can be drawn from a Gaussian distribution $\sim \mathcal{N}(0,I)$. However in other cases, where the input has been scaled differently, or has clearly defined input space boundaries, it is important to confirm the data distribution and input space boundaries, and design the sampling method such that samples conform to the actual input distribution as much as possible. The input distribution and input space bounds may be defined using the available validation or test set or based upon some prior knowledge. For example, for image input, the value for each pixel is bounded from 0 – 1, and each pixel position may have different value mean and variance. A random sampling strategy in this case could be to draw each pixel from a separate Gaussian distribution with mean and variance computed from the test set, and clipping the value between 0 to 1.

\par
\subsubsection{Generator model} \label{generator_model}
Adversarially trained generator models (GANs) are used to generate realistic synthetic data with the same statistics as the training set in a wide range of applications. Common applications include image generation, 3D model generation, audio generation, and other signal generation for purposes such as machinery fault diagnosis \cite{Aggarwal2021, Gui2023, Lou2022}.
\par
Generator model for generating synthetic data was proposed for data-free knowledge distillation for regression tasks by \cite{Kang2021}, follows similar methods in classification tasks \cite{Micaelli2019}. In this method, a generator model $G_{\phi}$ parameterized by $\phi$ is trained to output samples that would result in a large difference between the student and teacher model’s predictions. This generator model is trained in an adversarial manner against the student model during the distillation process by optimizing the generator loss function in equation \ref{eq:2}. 

\begin{equation} \label{eq:2}
L_G(z) = \mathbb{E}_{x_g ~ G_{\phi}(z)}[-(T(x_g)-S_{\theta}(x_g))^2]
\end{equation}

\begin{figure}
\centering
\includegraphics[width=\textwidth]{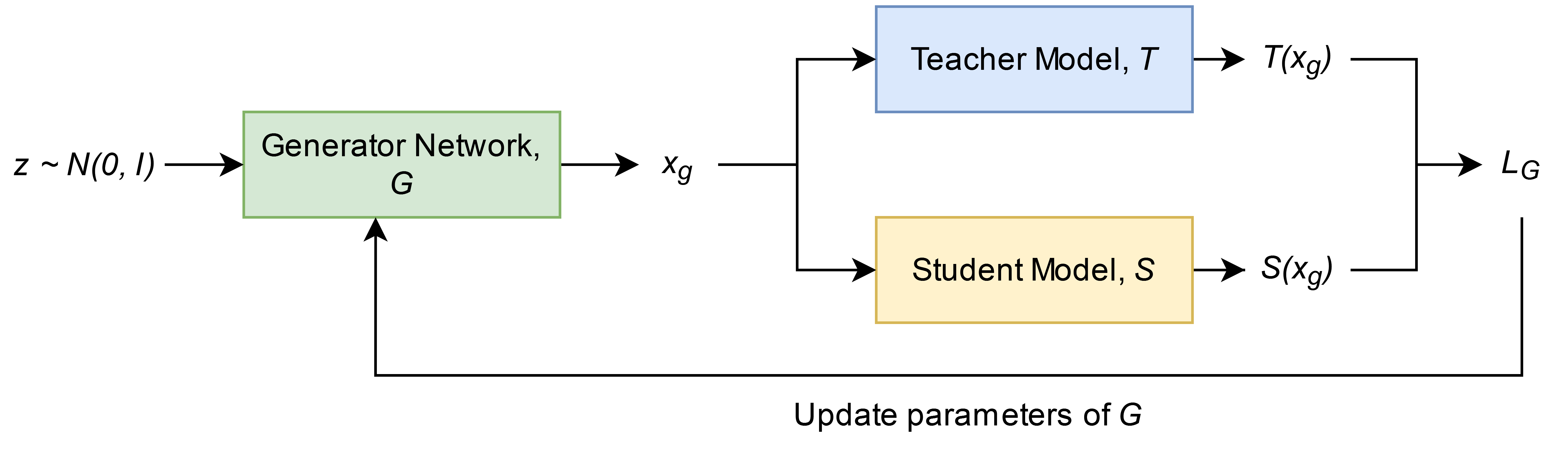}
\caption{\label{fig:fig2} Data-free distillation with generator method}
\end{figure}

The student is trained using the student loss to minimize the difference between teacher and its own predictions, the two opposing learning objectives are trained in a sequential adversarial manner, and the student model is able to learn to match the predictions of the teacher model as training continues. This process is illustrated in Figure \ref{fig:fig2}.

In practice, regularization terms may be added to the generator loss to prevent complete deviation from underlying data distribution, for e.g. by adding the square of $L_2$-norm of $x_g$ and $S_{\theta}(x_g)$, yielding:

\begin{equation} \label{eq:3}
L_G(z) = \mathbb{E}_{x_g ~ G_{\phi}(z)}[-(T(x_g)-S_{\theta}(x_g))^2 + \beta \|x_g\|^2 + \gamma S_{\theta}(x_g)^2]
\end{equation}

\subsubsection{Direct optimization from random samples}

The generator model approach attempts to train the generative model $G_{\theta}$ to approximate the inverse function of the student loss implicitly, where the generative model predicts $x$ given the objective of high student loss. It is not immediately clear whether the generative model is able to learn this inverse function easily. This is because generative adversarial models have been found to be difficult to train well empirically due to many reasons \cite{Arjovsky2017, Lucic2018, Salimans2016, Saxena2021}. Many design choices such as the number of training steps of the generator model vis-à-vis training steps of the student model and generator model architecture have large impacts on training stability and model performance and are difficult or tedious to tune. They often also suffer from problems such as mode collapse, and vanished gradients that make the generated synthetic data unsuitable for training the student model.
\par
Since the goal of the generative model approach is to generate samples that maximize the student loss, it is more straightforward to maximize the student loss directly, as formulated below.

\[
\max_{x_g}\ (T(x_g) - S_{\theta}(x_g))^2
\]

Or following conventions:

\begin{equation} \label{eq:4}
\min_{x_g}\ -(T(x_g) - S_{\theta}(x_g))^2
\end{equation}

In practice, following the generator method, we may add regularization terms as well, such as in equation \ref{eq:5}.

\begin{equation} \label{eq:5}
\min_{x_g} -(T(x_g) - S_{\theta}(x_g))^2 + \beta \|x_g\|^2 + \gamma (x_g)^2
\end{equation}

It is later shown in \ref{subsection_mnist} and \ref{subsection_prot} that the methodology is very flexible, and any arbitrary loss function may be used to incorporate loss terms designed to capture important properties of the data.
\par
\begin{figure}
\centering
\includegraphics[width=\textwidth]{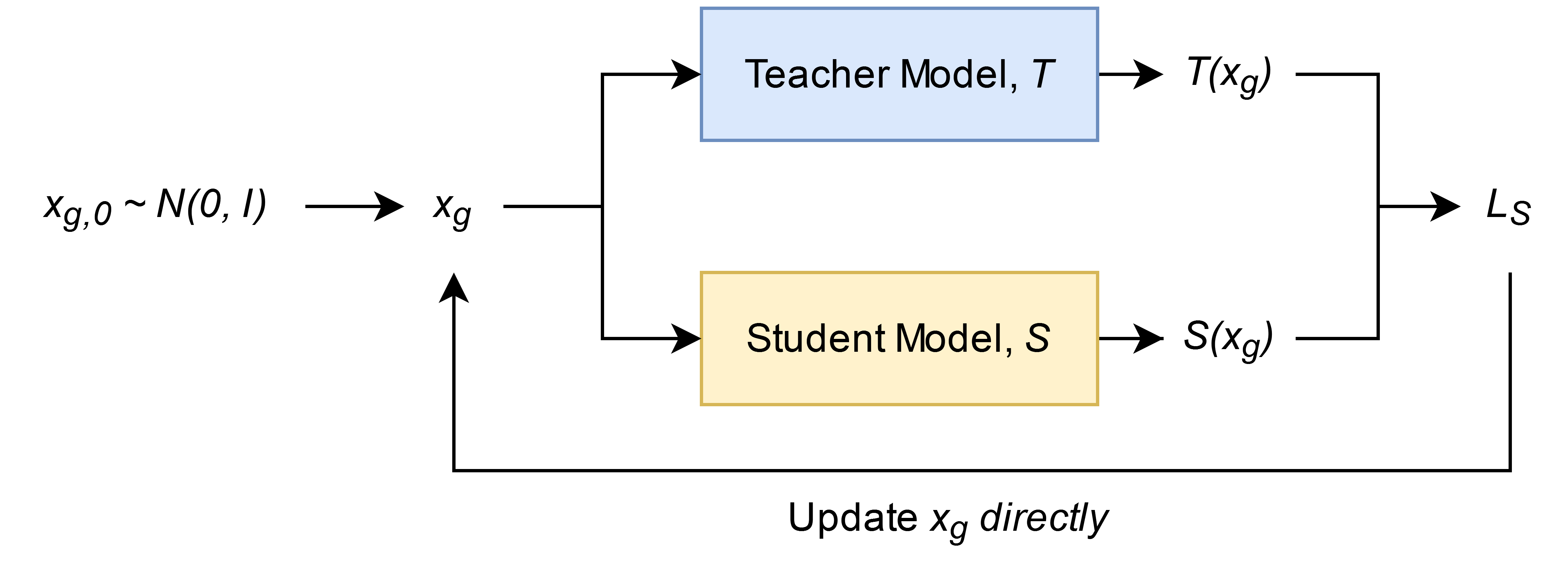}
\caption{\label{fig:fig3} Data-free model distillation with direct optimization method}
\end{figure}

This minimization can be done through various optimization algorithms. If both the student and teacher models are differentiable, gradient descent can be used. Black box metaheuristic optimization methods such as genetic algorithms and simulated annealing may also be used, especially if the teacher model gradients are unavailable. The method is illustrated in Figure \ref{fig:fig3}.

When using direct optimization of the student loss with gradient descent, it is possible to derive theoretical guarantees for (a) generating samples that are better than random sample and (b) generating samples that are bounded in their deviation away from underlying distribution. 
\par
The gradient descent updates as such:

\begin{equation} \label{eq:6}
x_{g, t+1} = x_{g, t} + \eta \frac{\partial}{\partial x_{g,t}}[T(x_g) - S_{\theta}(x_g)]^2
\end{equation}

Assuming the neural networks are locally smooth (Lipschitz continuous), given some sufficiently small learning rate $\eta$, $x_{g,t+1}$ always improves upon $x_{g,t}$  fulfilling guarantee (a). Given some learning rate $\eta$ and number of gradient descent steps $t_{max}$, $x_{g,t+1}$ deviates from $x_{g,0}$ randomly sampled from underlying distribution by an arbitrary bound, fulfilling guarantee (b). Proof for guarantee (a) is provided in \cite{Boyd2009} p.466 and proof for guarantee (b) is provided in the \hyperref[supplementary]{supplementary materials}.
\par
It is not obvious to us that the generator model method can fulfil guarantee (a) because $x_g$ is generated from Gaussian noise $z$ of an arbitrary dimension and is not related to random samples in input space; and to fulfil guarantee (b), a bound on the deviation of $x_g$ from 0 exist only if a regularization term is applied to $x_g$. The proof for bound on magnitude of $x_g$ for generator method with $L_2$ regularization is provided in the \hyperref[supplementary]{supplementary materials}.
\par
\subsubsection{Proposed method for knowledge distillation}

The proposed data-free knowledge distillation method generates training data $x_g$ through direct optimization of student loss with gradient descent. In the synthetic data generation step, assuming inputs are standardized, a batch of random samples are drawn from a Gaussian distribution $\sim \mathcal{N}(0,1)$. Gradient descent is used to perturb these random samples to the direction of maximizing their student loss values, obtaining $x_g$. In the student training step, the student weights are updated to minimize the student loss with respect to the synthetic data $x_g$.
\par
Following the methods proposed in \cite{Kang2021}, generated data is also supplemented with random samples $x_p$ drawn from Gaussian distribution $\sim \mathcal{N}(0,1)$. The sample weights for the generated samples $x_g$ and random samples $x_p$ are controlled by a factor $\alpha$, which can be a fixed value or follow a schedule based on the training epoch.

\begin{equation} \label{eq:7}
L_S = \alpha L_S(x_g) + (1-\alpha) L_S(x_p)
\end{equation}

Setting $\alpha$ to 0 is equivalent to the random sampling strategy. Setting $\alpha$ to 1 is a pure generative sampling strategy. Note that for both edge cases, since the loss of only 1 set of samples contributes to the training, the number of training samples in each epoch needs to be doubled for a fair comparison with cases where $\alpha$ is between 0 and 1. We investigate a decreasing alpha schedule as well as a pure $x_g$ training strategy in the experiments.
\par
\begin{figure}
\centering
\includegraphics[width=0.6\textwidth]{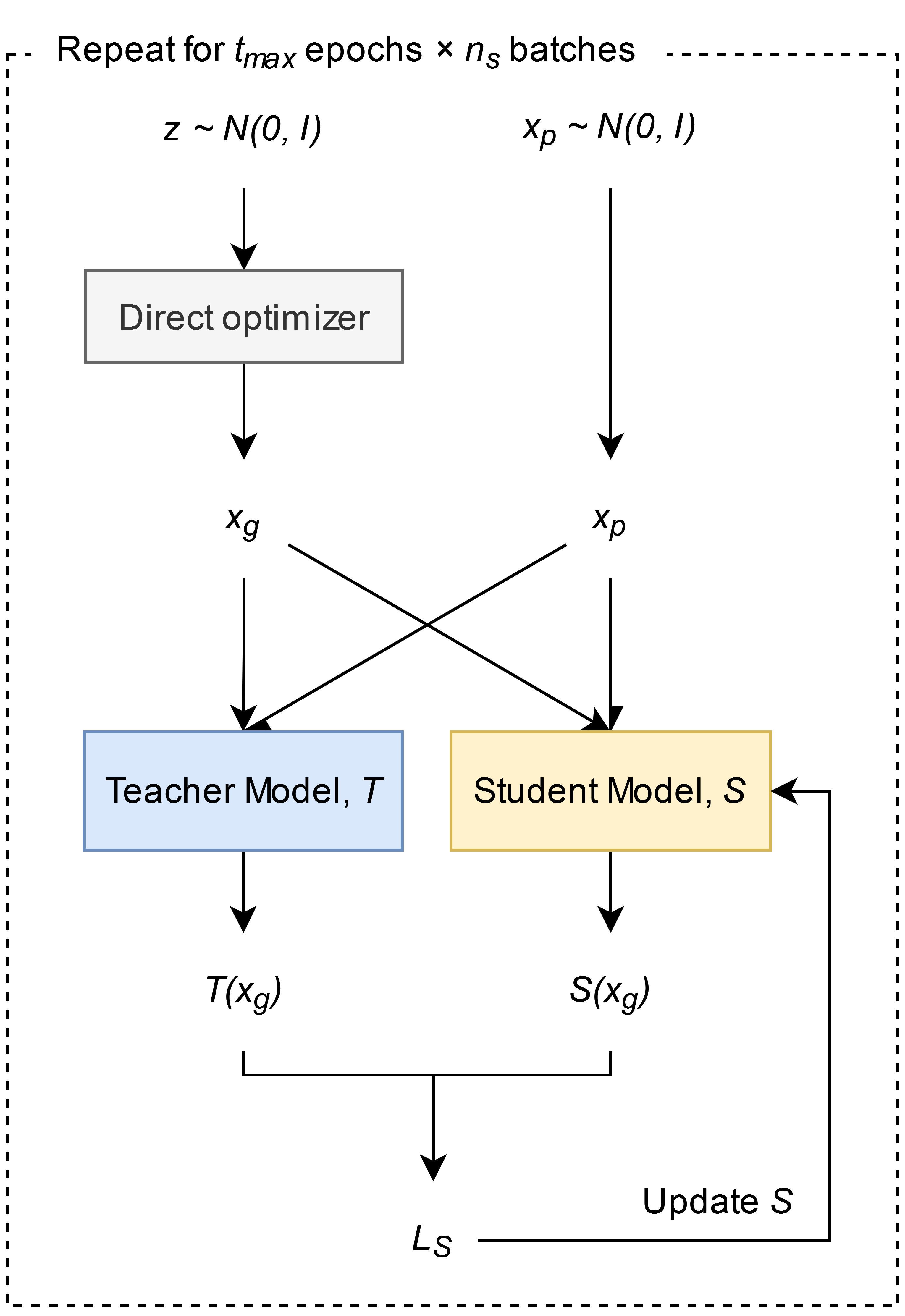}
\caption{\label{fig:fig4} Proposed data-free knowledge distillation method}
\end{figure}

The proposed method is illustrated in the block diagram in Figure \ref{fig:fig4}.

More formally, the training procedures are described in algorithm 1 \& 2. In the main procedure \textbf{\texttt{Data-free model distillation}} where the data distillation training happens, the number of training epochs for the student model is defined as $t_{max}$, and the number of batches per epoch is defined as $n_s$. In the sub-procedure \textbf{\texttt{Optimize}}, where direct optimization to generate synthetic data is done via gradient descent, the number of gradient descent steps is defined as $\tau_{max}$.

\SetKwInput{KwInput}{Input}                
\SetKwInput{KwOutput}{Output}              

\begin{algorithm}[!ht]
\DontPrintSemicolon
  
  \KwInput{teacher model, $T$}
  \KwOutput{student model, $S_{\theta}$}
  \For{$t$=1 to $t_{max}$}
  {
      \For{1 to $n_s$}
      {
        $z \sim \mathcal{N}(0, I)$ \;
        $x_g \gets$ \textbf{Optimize}$(z)$ \;
        $x_p \sim \mathcal{N}(0, I)$ \;
        $L \gets \alpha L_S(x_g) + (1-\alpha)L_S(x_p)$ \;
        Update $S_{\theta}$ with gradient descent w.r.t. $L$ \;
      }
  }

\caption{Main procedure: \textbf{Data-free model distillation}}
\end{algorithm}

\SetKwInput{KwInput}{Input}                
\SetKwInput{KwOutput}{Output}              

\begin{algorithm}[!ht]
\DontPrintSemicolon
  
  \KwInput{$z$}
  \KwOutput{$x_g$}
  $x_g \gets z$ \;
  \For{$\tau$=1 to $\tau_{max}$}
  {
      $L_S \gets -(T(x_g) - S_{\theta}(x_g))^2 + \beta \|x_g\|^2 + \gamma (x_g)^2$ \;
      $x_g \gets x_{g} - \eta \frac{\partial}{\partial x_{g}}L_S$ \;
  }

\caption{Sub-procedure: \textbf{Optimize}}
\end{algorithm}

\subsection{Regression datasets for experiments}
To facilitate comparison with the previous work by \cite{Kang2021}, the experiments were conducted on the same datasets. These 7 datasets are regression problem sets available from UCI machine learning repository \cite{Dheeru2019} and KEEL dataset repository \cite{AlcalaFdez2010}. ‘longitude’ was selected as the output variable for Indoorloc. Details of the datasets are provided in the Table \ref{tab:table1}.

\begin{table}[]
\centering
\caption{List of regression datasets}
\label{tab:table1}
\begin{tabular}{lll}
\hline
\textbf{Dataset} & \textbf{Number of features} & \textbf{Number of samples} \\ \hline
Compactiv        & 21                          & 8192                       \\
Cpusmall         & 12                          & 8192                       \\
CTScan           & 384                         & 53500                      \\
Indoorloc        & 520                         & 19337                      \\
Mv               & 10                          & 40768                      \\
Pole             & 26                          & 14998                      \\
Puma32h          & 32                          & 8192                       \\ \hline
\end{tabular}
\end{table}

The data are split into training and test set. The training set consists of 5000 samples for each dataset. 10\% of the remainder samples are placed into the validation set, and the remaining 90\% is the test set. The validation set is used to periodically evaluate the training of the student model.
\par
For data processing step, all values were standardized to a mean of 0 and a standard deviation of 1. Two processing workflows were tested where the scaling factors were calculated for the training set only and then applied to the test set, and where the scaling was done on the whole dataset prior to splitting of training and testing data. No significant differences were observed for both workflows, with differences in the RMSE of teacher models differing by less than 1.5\%. Therefore, the second workflow was used for the results for simplicity.

\subsection{Experiment setup for regression datasets}

To facilitate comparison, we used the same experiment setup for the neural networks as was used in \cite{Kang2021}. The teacher model is a fully connected feed forward network containing 1 hidden layer of 500 units with Tanh activation function. The student model is also a fully connected feed forward network containing 1 hidden layer of either 25, or 50 units with Tanh activation function.
\par
The teacher model is trained with the training data, while student models are trained without access to any real data from the training set. RMSProp optimizer is used for gradient descent, with a learning rate of $10^{-3}$ and weight decay regularization of $10^{-5}$. Batch size $m$ is set to be 50, and the number of batches in each epoch, $n_s$ is set to be 10. $\beta$ and $\gamma$ are selected to be $10^{-5}$. The number of epochs is selected as 2000. Models that performed the best on the validation loss was used to evaluate on the test set. For the direct optimization method to generate synthetic data, RMSProp optimizer with a learning rate of $10^{-1}$, and 2 epochs were used, how these two hyperparameters were selected are elaborated in the results section \ref{Properties}.

\subsection{Experiments on MNIST dataset} \label{subsection_mnist}

To further test the applicability of our method on different types of inputs, and on deeper and more complex neural network architectures, we designed an experiment for data-free knowledge distillation for regression on the MNIST handwritten digits dataset.
\par
The MNIST dataset is originally intended to be used for classification task, following the method presented in \cite{Wang2020}, we adapt it for regression task by making the neural network to predict a continuous number that represent the class value of the digit label of the input image. The performance of the model is measured in mean absolute error (MAE) between the predicted value and the actual value of the digit. For e.g. for perfect performance, the model should predict a value of 3.0 for an image with the handwritten digit 3. A prediction of 2.9 will result in a MAE of 0.1.
\par
The input image in MNIST is a single channel image of size 28 by 28 pixels, each pixel taking a value between 0 – 1. The mean $\mu$ and standard deviation $\sigma$ of each pixel position is calculated for the entire dataset and is used to generate random datapoints with a normal distribution $\mathcal{N}(\mu, \sigma)$ clipped between 0 – 1. This is done to achieve as much as possible a similar distribution as real data.
\par
As proposed by \cite{Wang2020}, we used a multi-layer convolutional neural network with the architecture specified in Table \ref{tab:table2}. The teacher and student network follow the same architecture, except that the number of filters, $f$ for each convolutional layer in the teacher network is higher than that in the student network. $f$ is chosen to be 10 for the teacher network and 5 for the student network. Log hyperbolic cosine (Log-Cosh) loss was used instead of mean squared error as the loss function to improve training following method proposed by \cite{Wang2020}.

\begin{table}[]
\centering
\caption{Architecture of neural network for MNIST regression}
\label{tab:table2}
\begin{tabular}{lll}
\hline
\textbf{Name}     & \textbf{Filters/units} & \textbf{Activation function} \\ \hline
Conv2D-1          & 3 x 3 x f              & ReLU                         \\
Conv2D-2          & 3 x 3 x 2f             & softplus                     \\
Maxpool2D-1       & 2 x 2                  &                              \\
Conv2D-3          & 3 x 3 x 4f             & softplus                     \\
Maxpool2D-2       & 2 x 2                  &                              \\
Flatten           &                        &                              \\
Fully connected-1 & 500                    & softplus                     \\
Dropout-1 (0.5)   &                        &                              \\
Fully connected-1 & 100                    & softplus                     \\
Dropout-2 (0.25)  &                        &                              \\
Fully connected-1 & 20                     & softplus                     \\ 
Fully connected-1 & 1                      & softplus                     \\ \hline
\end{tabular}
\end{table}

Due to the different nature of the input, which are images rather than standardized tabular data in the regression datasets, and the output which are natural number, we designed a different loss function for generating synthetic data. This loss function differs from equations \ref{eq:3} and \ref{eq:5} by replacing the mean-squared error loss with Log-Cosh loss and by changing the regularization terms to better capture the distribution of real data. Firstly, instead of penalizing the $L_2$ norm of $x_g$, we penalize the $L_1$ norm of $x_g$ because the handwritten digits image tends to be sparse. Secondly, instead of penalizing the student prediction on $x_g$, we randomly sample a whole number from 0 – 9 and penalize the distance of the teacher’s prediction to the random whole number. The purpose of this penalty is to allow the synthetically generated sample $x_g$ to match more closely with the actual data distribution, as real data should generally not be predicted too far away from whole number by the teacher model for this task. This strategy may also be helpful for other regression tasks where the output lies in a fixed set of discrete values.

\[
y_{rand} \sim \{n \in \mathbb{Z} : 0 \leq n \leq 9\}
\]

\begin{equation} \label{eq:8}
L_G(x_g) = -\epsilon\ log[cosh(T(x_g) - S_{\theta}(x_g))] + \beta \lvert x_g \lvert + \gamma (T(x_g) - y_{rand})^2
\end{equation}

To train the student model, RMSProp optimizer was used for gradient descent, with a learning rate of $10^{-3}$ and weight decay regularization of $10^{-5}$. Batch size $m$ is set to be 50, and the number of batches in each epoch, $n_s$ is set to be 10. The number of epochs is selected as 1000. 
\par
For the direct optimization method to generate synthetic data, RMSProp optimizer with a learning rate of $10^{-3}$, and 20 epochs were used. For the generator network, the number of rounds for training the generator per epoch was also set to 20. 

\subsection{Case study on protein solubility prediction} \label{subsection_prot}

A bioinformatics problem, predicting continuous protein solubility value with the constituent amino acids \cite{Han2019}, was used as a case study to test the effectiveness of data-free knowledge distillation for regression on a real-world scientific problem. Predicting continuous solubility value is useful for \textit{in-silico} screening and design of proteins for industrial applications \cite{Han2020}. 
\par
We also want to test how the method can be used when the gradients of the teacher model are not available. For example, many bioinformatics tools such as protein solubility prediction are hosted on servers that allow users to query proteins and obtain predictions. However, both the model and data used to train the model are not available to the user. To recreate the model, data-free knowledge distillation without gradient access to the teacher model is required. If gradient information of the teacher model is unavailable, it is not possible to train the generative network as described in \ref{generator_model} directly. However, for direct optimization, it is possible to use metaheuristics optimization that does not require gradients instead of gradient descent. 
\par
The dataset used contains 3148 proteins with solubility represented as a continuous value between 0 – 1 from the eSol database \cite{Niwa2009}. The input features are the proportion of each of the 20 amino acids within the protein sequence. 2500 proteins are selected for the training set, and the remaining as test set. The teacher model used is a support vector machine, which represents the black-box teacher model that contains no gradient information and only output prediction value is available.
\par
As in the MNIST example, we introduce diversity in the predicted value by the teacher model on $x_g$ with a penalty term on distance away from a random $y$ value sampled for every batch.

\[
y_{rand} \sim \{n \in \mathbb{R}^+ : 0 \leq n \leq 1\}
\]

\begin{equation} \label{eq:9}
L_G(x_g) = -\epsilon\ (T(x_g) - S_{\theta}(x_g))^2 + (1-\epsilon)(T(x_g) - y_{rand})^2
\end{equation}

The student model is made up of a fully connected Gaussian kernel radial basis function layer with output size of 100, followed by a fully connected linear layer that outputs the prediction. For training the student models in both baseline and direct optimization method, RMSProp optimizer is used for gradient descent, with a learning rate of $10^{-3}$ and weight decay regularization of $10^{-6}$. Batch size $m$ is set to be 50 with decreasing $\alpha$ schedule. 
\par
Random sampling was used for training baseline model and providing initial points for direct optimization method. The mean $\mu$ and standard deviation $\sigma$ of each amino acid feature is calculated from the training dataset and is used to generate random datapoints with a normal distribution $\mathcal{N}(\mu, \sigma)$ clipped between 0 – 1. The feature values are then normalized such that the value sums to 1. This is done as the features which are proportion of each of the 20 amino acids within the protein sequence must sum to 1. For the direct optimization method to generate synthetic data, differential evolution algorithm \cite{Storn1997} with 25 iterations of \textit{best2bin} strategy was used, with initial points generated with the random sampling just described. 

\section{Results}
\subsection{Properties of synthetic data generated} \label{Properties}

We first investigate the properties of the synthetic data generated by various methods, namely the student loss value of the synthetic data, and the distribution of the synthetic data.

\subsubsection{Student loss value of synthetic data}

Intuitively, the goal of the synthetic data generation process is to generate data that gives large differences in student and teacher prediction (i.e. student loss $L_S$ in equation \ref{eq:1}) in the hope that by learning to correct these large mistakes, the student model is able to learn faster and better mimic the outputs of the teacher model.
\par
\begin{figure}[!ht]
\centering
\includegraphics[width=\textwidth]{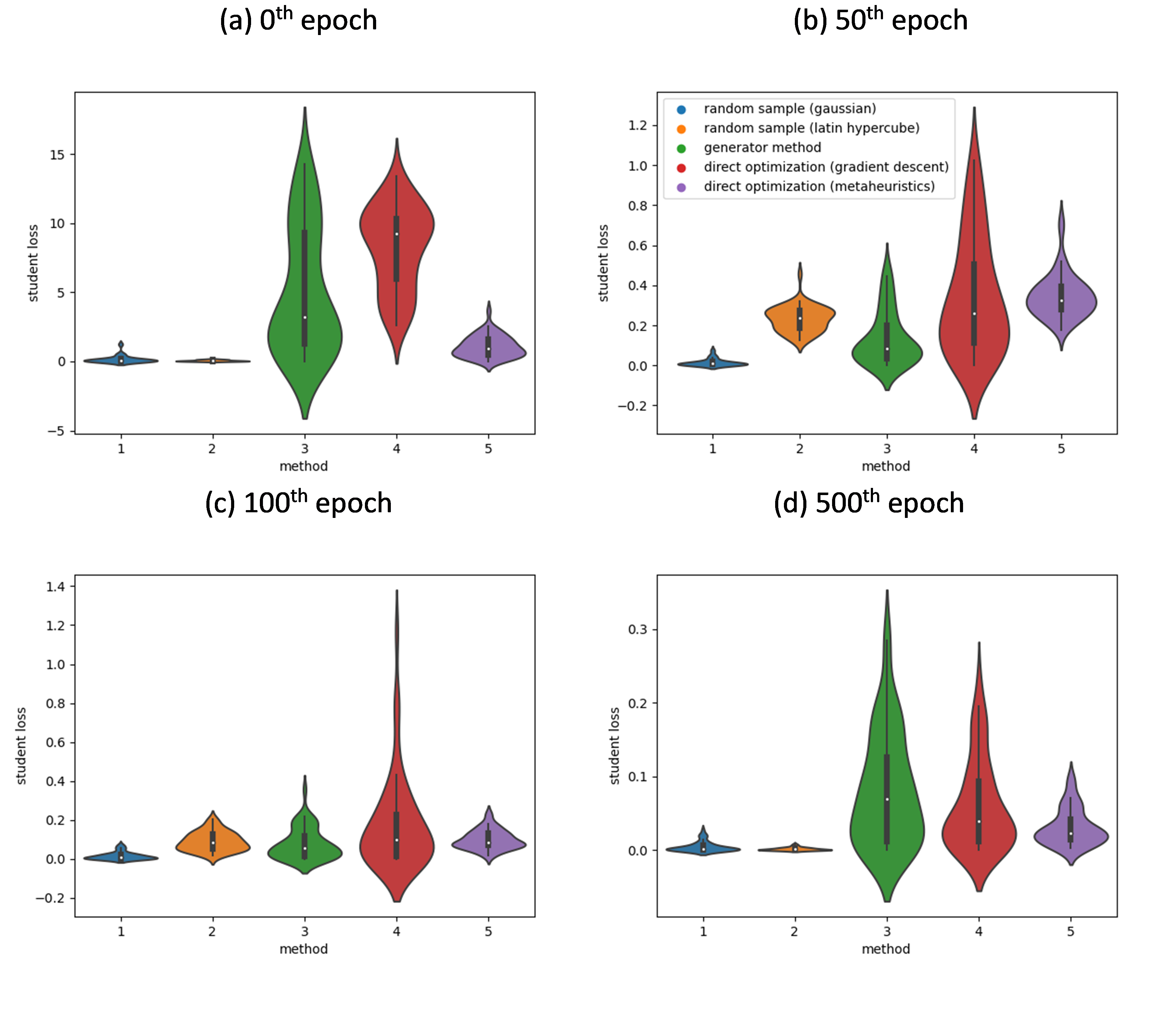}
\caption{\label{fig:fig5} Violin plot of student loss of synthetic data at different epochs}
\end{figure}

To verify the actual behavior of the various methods at achieving this goal, we compare the student loss values of synthetic data generated by the various methods at different stages of training a student model with random samples: when student is first randomly initialized at 0\textsuperscript{th} epoch, during the middle stage of training at the 50\textsuperscript{th} and 100\textsuperscript{th} epoch, and when the student model has converged at the 500\textsuperscript{th} epoch. The results shown in Figure \ref{fig:fig5} are for Indoorloc dataset.
\par
As expected, it can be observed that the synthetic data generated by the generator method and the direct optimization method have higher student loss than random Gaussian samples at all stages of training. Compared to the direct optimization method, the generator method tends to generate data with smaller loss at the early stages of training, and larger loss at later stages of training.
\par
Directly optimizing with metaheuristics algorithms, in this case differential evolution is also capable of generating synthetic data with high loss. However, the running speed of metaheuristics algorithms is much slower than gradient descent and is not ideal practically unless gradient information is unavailable. 

\subsubsection{Distribution of synthetic data}

Synthetic data generated should reasonably overlap with the underlying distribution. Out of distribution data generated may either be not useful or even detrimental to model performance on test data. Ideally the synthetic data generated should also be well spread out from each other rather than clustered closely together to allow for better coverage of the data distribution.
\par
\begin{figure}[!ht]
\centering
\includegraphics[width=\textwidth]{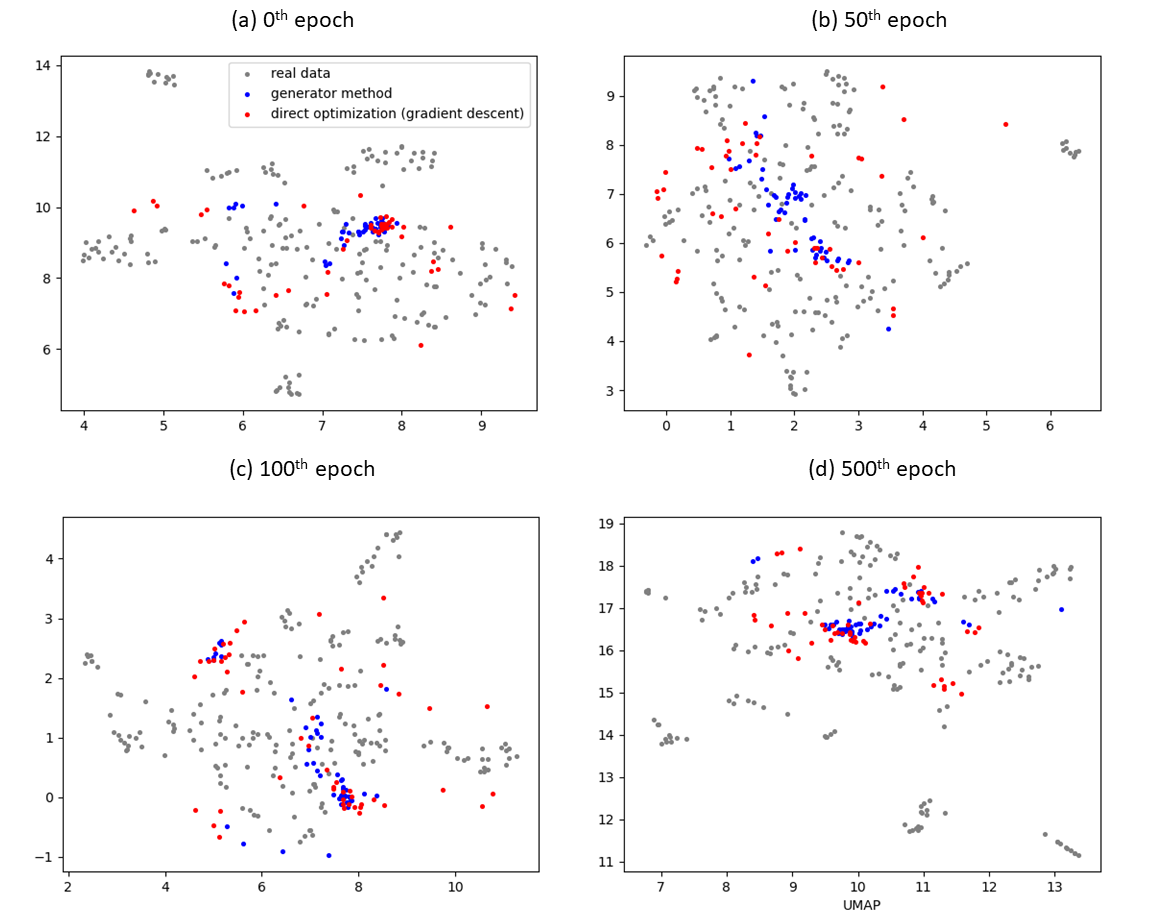}
\caption{\label{fig:fig6} Distribution of synthetic datapoints at different epochs}
\end{figure}

To verify the actual behavior of the generator and direct optimization methods at achieving this goal, we visualize the distribution of synthetic datapoints generated by the generator method and direct optimization (gradient descent) method at different stages of training using plots of 2D UMAP (Uniform Manifold Approximation and Projections) shown in Figure \ref{fig:fig6}.

\par
It is observed that the synthetic data generated by the generator approach tends to converge around one or two tight clusters, leaving the rest of the input space untouched. Even though the direct optimization approach also tends to have some datapoints concentrated at a few clusters, the rest of the datapoints tends to be much better spread out in the input space, while still maintaining similarity with real data. This suggests greater diversity of synthetic data generated with direct optimization should be helpful for training the student model. 
\par
It is also observed that at the later stage of training, many more datapoints generated by the generator method cluster at regions where there are no real datapoints compared to datapoints generated by the direct optimization method. This may explain the larger student loss for the generator method than direct optimization method at later stage of training. This suggests that the decreasing schedule for sample weights parameter $\alpha$ which controls how much the generated data, $x_g$ influence the training loss compared to random samples $x_p$, would likely play a much more important role when using the generator method. Because at later stage of training, $x_g$ generated by the generator method will likely deviate more from the underlying distribution and may lead to negative learning, which necessitates a smaller weight $\alpha$.
\par
We have experimented and found that direct optimizing for 2 steps with a step size of $10^{-1}$ leads to generating synthetic data that do not deviate much from the underlying distribution while still providing a substantially higher student loss than random samples. Hence these two hyperparameters were selected for the direct optimization method.

\subsection{Comparison of different methods for data-free distillation on regression datasets}

Table \ref{tab:table3} and Table \ref{tab:table4} shows the comparison of root mean squared error (RMSE) for 5 methods of data-free distillation using student size of 25 and 50 respectively. For the generator method and direct optimization method, both a decreasing $\alpha$ schedule and $\alpha$ value of 1 are tested. The $\alpha$ value of 1 means that the training uses the generated synthetic data $x_g$ entirely without any randomly sampled datapoints.
\par
Comparing the results for student model size of 25 and 50 hidden units, it is observed that with an increase in student model size, the RMSE is lower for all datasets due to the greater representation power of the student model. For most of the datasets tested, the direct optimization method achieves the lowest RMSE and most closely matches the performance of the teacher model. Compared against random sampling, direct optimization with decreasing alpha achieves lower RMSE on 6 out of 7 datasets. Compared against generator method with decreasing $\alpha$, direct optimization with decreasing $\alpha$ achieves lower RMSE on 5 out of 7 datasets.

\begin{table}[h!]
\centering
\caption{RMSE results achieved with different methods for student model size of 25}
\label{tab:table3}
\begin{tabular}{p{0.09\textwidth}p{0.13\textwidth}p{0.13\textwidth}p{0.13\textwidth}p{0.13\textwidth}p{0.13\textwidth}p{0.13\textwidth}}
\hline
\textbf{Dataset} & \textbf{Teacher Model}                                          & \textbf{Random Sampling}                                        & \textbf{Generator; decreasing $\alpha$}                         & \textbf{Generator; $\alpha$ = 1}                                & \textbf{Direct optimizer; decreasing $\alpha$}                  & \textbf{Direct optimizer; $\alpha$ = 1}                         \\ \hline
Compactv         & \begin{tabular}[c]{@{}l@{}}0.1441\\ ± 0.0039\end{tabular} & \begin{tabular}[c]{@{}l@{}}0.1588 \\ ± 0.0050\end{tabular}         & \begin{tabular}[c]{@{}l@{}}0.1606\\ ± 0.0061\end{tabular}          & \begin{tabular}[c]{@{}l@{}}0.1693\\ ± 0.0069\end{tabular} & \textbf{\begin{tabular}[c]{@{}l@{}}0.1562\\ ± 0.0043\end{tabular}} & \begin{tabular}[c]{@{}l@{}}0.1599\\ ± 0.0067\end{tabular}          \\ \\
Cpusmall         & \begin{tabular}[c]{@{}l@{}}0.1672\\ ± 0.0031\end{tabular} & \begin{tabular}[c]{@{}l@{}}0.1840\\ ± 0.0065\end{tabular}          & \begin{tabular}[c]{@{}l@{}}0.1875\\ ± 0.0070\end{tabular}          & \begin{tabular}[c]{@{}l@{}}0.1918\\ ± 0.0101\end{tabular} & \textbf{\begin{tabular}[c]{@{}l@{}}0.1817\\ ± 0.0042\end{tabular}} & \begin{tabular}[c]{@{}l@{}}0.1822\\ ± 0.0048\end{tabular}       \\ \\
CTScan           & \begin{tabular}[c]{@{}l@{}}0.1058\\ ± 0.0060\end{tabular} & \begin{tabular}[c]{@{}l@{}}0.2248\\ ± 0.0170\end{tabular}          & \begin{tabular}[c]{@{}l@{}}0.1601\\ ± 0.0044\end{tabular}          & \begin{tabular}[c]{@{}l@{}}0.2091\\ ± 0.0090\end{tabular} & \begin{tabular}[c]{@{}l@{}}0.1649\\ ± 0.0058\end{tabular}          & \begin{tabular}[c]{@{}l@{}}0.1593\\ ± 0.0054\end{tabular}          \\ \\
Indoorloc        & \begin{tabular}[c]{@{}l@{}}0.0847\\ ± 0.0018\end{tabular} & \begin{tabular}[c]{@{}l@{}}0.105\\ ± 0.0051\end{tabular}           & \begin{tabular}[c]{@{}l@{}}0.1034\\ ± 0.0034\end{tabular}          & \begin{tabular}[c]{@{}l@{}}0.1629\\ ± 0.0134\end{tabular} & \textbf{\begin{tabular}[c]{@{}l@{}}0.0944\\ ± 0.0015\end{tabular}} & \begin{tabular}[c]{@{}l@{}}0.0957\\ ± 0.0035\end{tabular}          \\ \\
Mv               & \begin{tabular}[c]{@{}l@{}}0.0236\\ ± 0.0022\end{tabular} & \textbf{\begin{tabular}[c]{@{}l@{}}0.0250\\ ± 0.0019\end{tabular}} & \begin{tabular}[c]{@{}l@{}}0.0255\\ ± 0.0016\end{tabular}          & \begin{tabular}[c]{@{}l@{}}0.0428\\ ± 0.0045\end{tabular} & \begin{tabular}[c]{@{}l@{}}0.0252\\ ± 0.0016\end{tabular}          & \begin{tabular}[c]{@{}l@{}}0.0284\\ ± 0.0017\end{tabular}          \\ \\
Pole             & \begin{tabular}[c]{@{}l@{}}0.1549\\ ± 0.0064\end{tabular} & \begin{tabular}[c]{@{}l@{}}0.2893\\ ± 0.0141\end{tabular}          & \textbf{\begin{tabular}[c]{@{}l@{}}0.2748\\ ± 0.0161\end{tabular}} & \begin{tabular}[c]{@{}l@{}}0.3484\\ ± 0.0304\end{tabular} & \begin{tabular}[c]{@{}l@{}}0.2836\\ ± 0.0198\end{tabular}          & \begin{tabular}[c]{@{}l@{}}0.3523\\ ± 0.0206\end{tabular}          \\ \\
Puma32h          & \begin{tabular}[c]{@{}l@{}}0.2589\\ ± 0.0055\end{tabular} & \begin{tabular}[c]{@{}l@{}}0.2474\\ ± 0.0043\end{tabular}          & \begin{tabular}[c]{@{}l@{}}0.2499\\ ± 0.0035\end{tabular}          & \begin{tabular}[c]{@{}l@{}}0.2686\\ ± 0.0091\end{tabular} & \begin{tabular}[c]{@{}l@{}}0.2464\\ ± 0.0034\end{tabular}          & \textbf{\begin{tabular}[c]{@{}l@{}}0.2460\\ ± 0.0034\end{tabular}} \\ \hline
\end{tabular} 
\end{table}

\begin{table}[h!]
\centering
\caption{RMSE results achieved with different methods for student model size of 50}
\label{tab:table4}
\begin{tabular}{p{0.09\textwidth}p{0.13\textwidth}p{0.13\textwidth}p{0.13\textwidth}p{0.13\textwidth}p{0.13\textwidth}p{0.13\textwidth}}
\hline
\textbf{Dataset} & \textbf{Teacher Model}                                          & \textbf{Random Sampling}                                        & \textbf{Generator; decreasing $\alpha$}                         & \textbf{Generator; $\alpha$ = 1}                                & \textbf{Direct optimizer; decreasing $\alpha$}                  & \textbf{Direct optimizer; $\alpha$ = 1}                         \\ \hline
Compactv         & \begin{tabular}[c]{@{}l@{}}0.1450\\ ± 0.0062\end{tabular} & \begin{tabular}[c]{@{}l@{}}0.15534\\ ± 0.0077\end{tabular} & \begin{tabular}[c]{@{}l@{}}0.1551\\ ± 0.0060\end{tabular}          & \begin{tabular}[c]{@{}l@{}}0.1837\\ ± 0.0124\end{tabular} & \textbf{\begin{tabular}[c]{@{}l@{}}0.1514\\ ± 0.0068\end{tabular}} & \begin{tabular}[c]{@{}l@{}}0.1531\\ ± 0.0066\end{tabular}          \\ \\
Cpusmall         & \begin{tabular}[c]{@{}l@{}}0.1663\\ ± 0.0037\end{tabular} & \begin{tabular}[c]{@{}l@{}}0.1760\\ ± 0.0043\end{tabular}  & \begin{tabular}[c]{@{}l@{}}0.1744\\ ± 0.0040\end{tabular}          & \begin{tabular}[c]{@{}l@{}}0.1842\\ ± 0.0079\end{tabular} & \textbf{\begin{tabular}[c]{@{}l@{}}0.1737\\ ± 0.0027\end{tabular}} & \textbf{\begin{tabular}[c]{@{}l@{}}0.1737\\ ± 0.0049\end{tabular}} \\ \\
CTScan           & \begin{tabular}[c]{@{}l@{}}0.1032\\ ± 0.0048\end{tabular} & \begin{tabular}[c]{@{}l@{}}0.1980\\ ± 0.0111\end{tabular}  & \begin{tabular}[c]{@{}l@{}}0.1458\\ ± 0.0058\end{tabular}          & \begin{tabular}[c]{@{}l@{}}0.2165\\ ± 0.0092\end{tabular} & \begin{tabular}[c]{@{}l@{}}0.1320\\ ± 0.0047\end{tabular}          & \begin{tabular}[c]{@{}l@{}}0.1316\\ ± 0.0050\end{tabular}          \\ \\
Indoorloc        & \begin{tabular}[c]{@{}l@{}}0.0844\\ ± 0.0039\end{tabular} & \begin{tabular}[c]{@{}l@{}}0.0965\\ ± 0.0043\end{tabular}  & \begin{tabular}[c]{@{}l@{}}0.1020\\ ± 0.0035\end{tabular}          & \begin{tabular}[c]{@{}l@{}}0.1549\\ ± 0.0076\end{tabular} & \textbf{\begin{tabular}[c]{@{}l@{}}0.0890\\ ± 0.0021\end{tabular}} & \begin{tabular}[c]{@{}l@{}}0.0913\\ ± 0.0027\end{tabular}          \\ \\
Mv               & \begin{tabular}[c]{@{}l@{}}0.0226\\ ± 0.0027\end{tabular} & \begin{tabular}[c]{@{}l@{}}0.0237\\ ± 0.0023\end{tabular}  & \textbf{\begin{tabular}[c]{@{}l@{}}0.0235\\ ± 0.0019\end{tabular}} & \begin{tabular}[c]{@{}l@{}}0.0441\\ ± 0.0059\end{tabular} & \textbf{\begin{tabular}[c]{@{}l@{}}0.0235\\ ± 0.0021\end{tabular}} & \begin{tabular}[c]{@{}l@{}}0.0271\\ ± 0.0022\end{tabular}          \\ \\
Pole             & \begin{tabular}[c]{@{}l@{}}0.1539\\ ± 0.0055\end{tabular} & \begin{tabular}[c]{@{}l@{}}0.2163\\ ± 0.0143\end{tabular}  & \textbf{\begin{tabular}[c]{@{}l@{}}0.1964\\ ± 0.0074\end{tabular}} & \begin{tabular}[c]{@{}l@{}}0.2094\\ ± 0.0059\end{tabular} & \begin{tabular}[c]{@{}l@{}}0.2092\\ ± 0.0114\end{tabular}          & \begin{tabular}[c]{@{}l@{}}0.2324\\ ± 0.0165\end{tabular}          \\ \\
Puma32h          & \begin{tabular}[c]{@{}l@{}}0.2625\\ ± 0.0057\end{tabular} & \begin{tabular}[c]{@{}l@{}}0.2521\\ ± 0.0035\end{tabular}  & \begin{tabular}[c]{@{}l@{}}0.2554\\ ± 0.0036\end{tabular}          & \begin{tabular}[c]{@{}l@{}}0.2639\\ ± 0.0123\end{tabular} & \begin{tabular}[c]{@{}l@{}}0.2518\\ ± 0.0057\end{tabular}          & \textbf{\begin{tabular}[c]{@{}l@{}}0.2495\\ ± 0.0045\end{tabular}} \\ \hline
\end{tabular} 
\end{table}

When $\alpha$ is set to 1, we observe a substantial increase in RMSE for the generator method. However, for the direct optimization method, setting $\alpha$ to 1 generally does not lead to much worse performance. This matches our hypothesis that a decreasing $\alpha$ schedule is much more important for the generator method as the synthetic datapoint generated tends to deviate more from the underlying distribution at a later stage of training. Compared with generator method with decreasing $\alpha$, direct optimization $\alpha = 1$ (i.e. training on $x_g$ only) achieves lower RMSE on 5 out of 7 datasets.
\par
\begin{figure}[ht!]
\centering
\includegraphics[width=\textwidth]{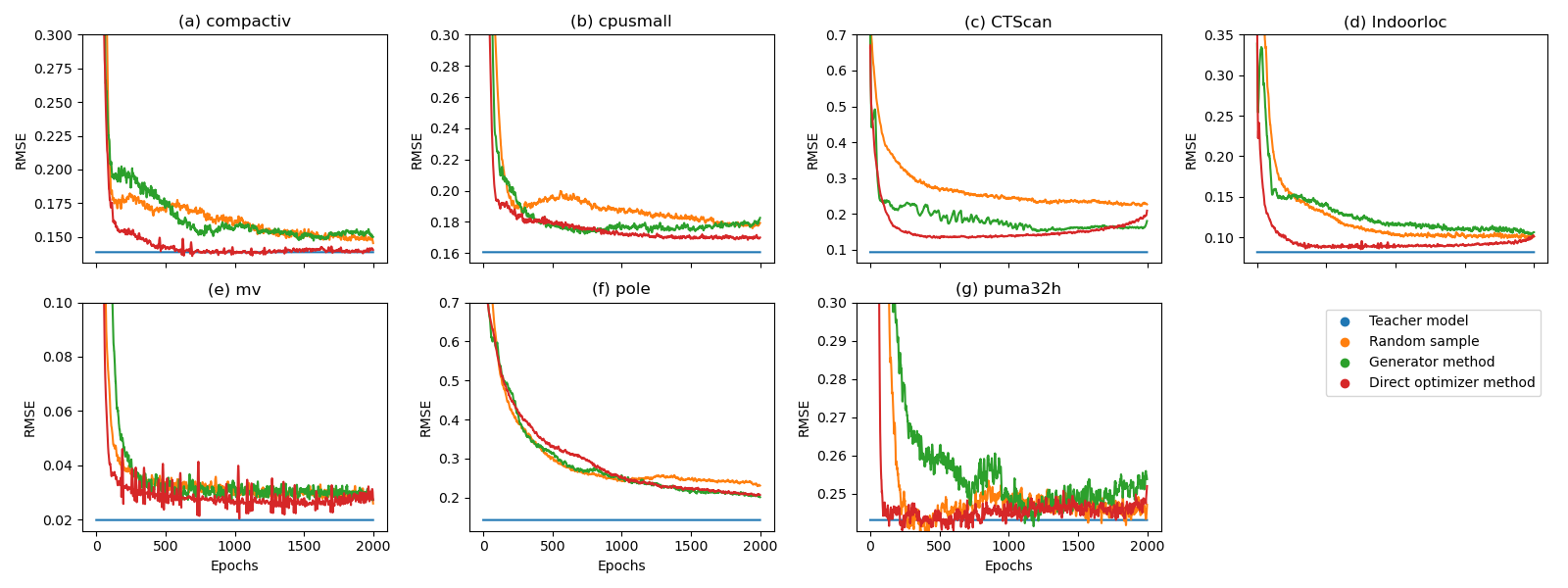}
\caption{\label{fig:fig7} RMSE on the validation set against training epochs}
\end{figure}

Figure \ref{fig:fig7} shows the RMSE on the validation set over the course of training of the student model of size 50. The direct optimization method shows a faster decrease in RMSE and a generally more stable learning behavior than the generator method. (Plot values have been smoothed with a Savitzky–Golay filter of window size of 15 epochs to reduce noise for better visualization)

\begin{figure}[ht!]
\centering
\includegraphics[width=\textwidth]{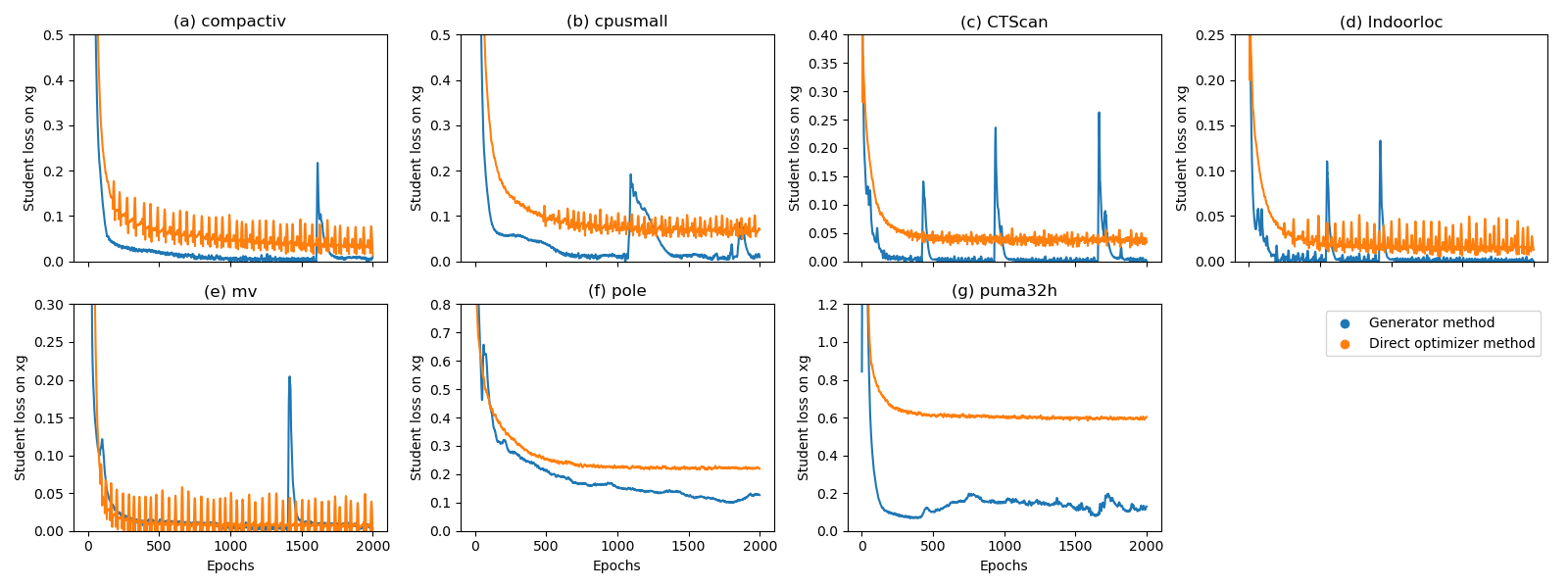}
\caption{\label{fig:fig8} Student loss on synthetic data $x_g$ against training epochs}
\end{figure}

We also examine the student loss on $x_g$ for the two models where $\alpha = 1$. As seen in Figure \ref{fig:fig8}, the generator method often produce unexpectedly large losses during training that could results in negative learning for the model, whereas the direct optimization method generally produces a stable and consistently decreasing loss.

\subsection{Comparison of different methods for data-free distillation on MNIST}

We experimented with different settings of $\beta$, $\gamma$ and $\epsilon$ weights for the various components in the loss function (equation \ref{eq:8}) and found that a low $\beta$ value ($10^{-6}$), high $\gamma$ value (set to 1), low $\epsilon$ value (set to $10^{-6}$) and provides good regularization that encourages synthetic data generated to be diverse and resemble real data distribution more closely.
\par
Table \ref{tab:table5} below shows the comparison of mean absolute error (MAE) as well as the RMSE achieved by training the student model with synthetic data sampled randomly, generated by the generator method and by the direct optimization method. MAE was used as the evaluation metric following the original study \cite{Wang2020}. Results are averaged over 5 runs. Note that the best performing random model that outputs a constant value of 4.5 would give a MAE of approximately 2.5 for a class balanced test set.

\begin{figure}[ht!]
\centering
\includegraphics[width=\textwidth]{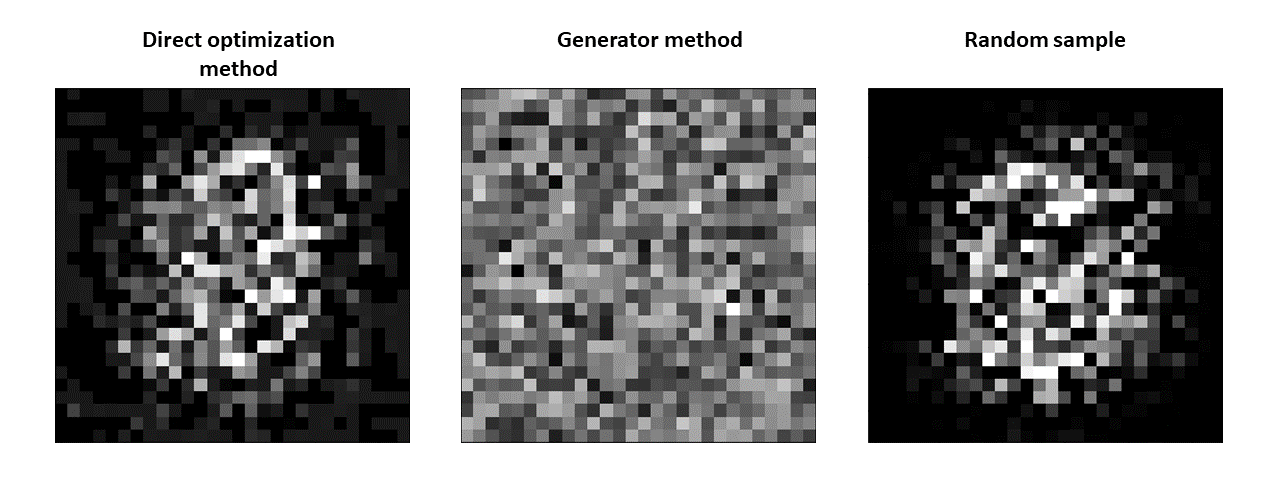}
\caption{\label{fig:fig9} Samples of a synthetic image generated by different methods}
\end{figure}
\begin{figure}[ht!]
\centering
\includegraphics[width=\textwidth]{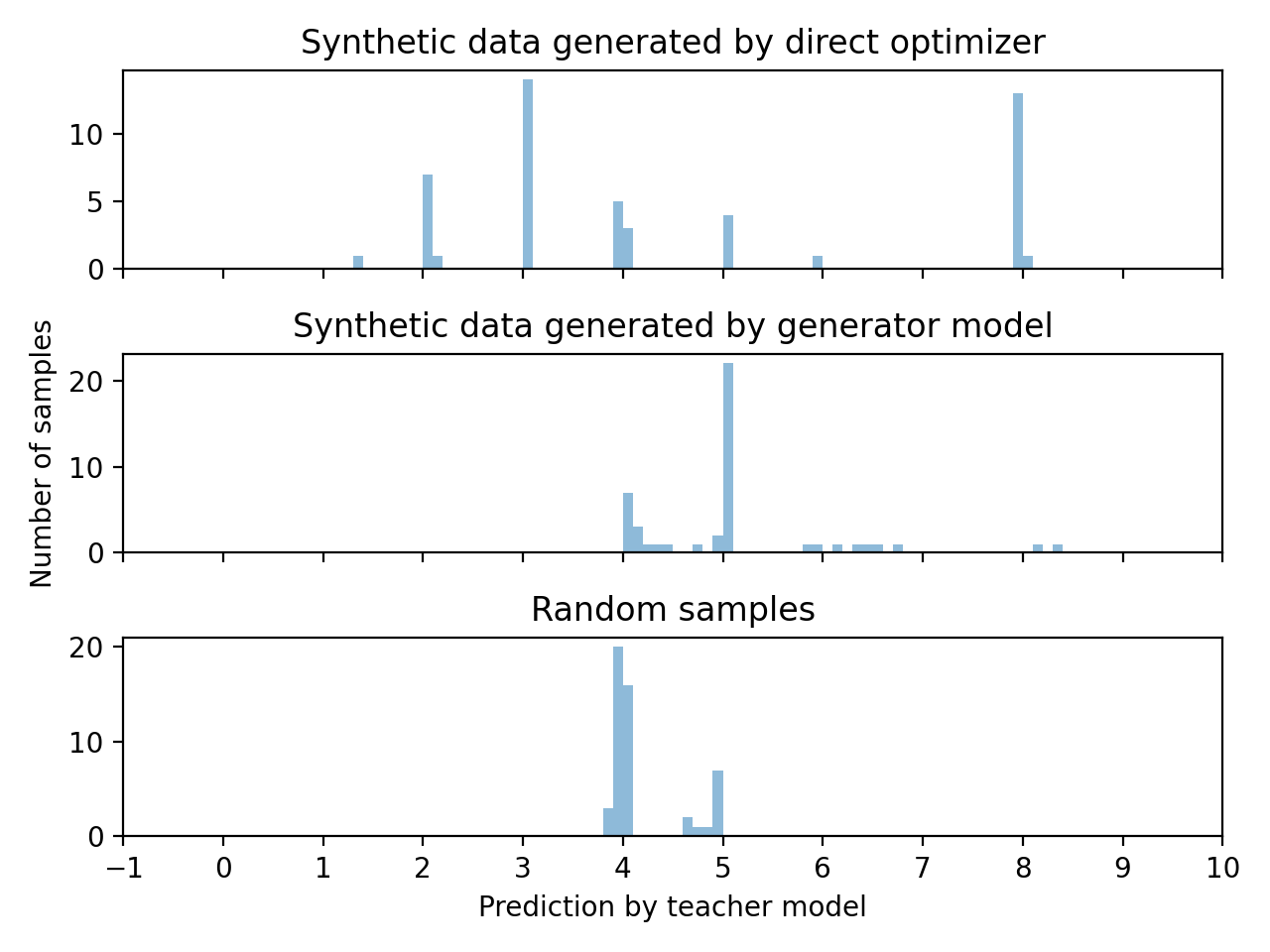}
\caption{\label{fig:fig10} Histograms of predicted values by teacher networks for synthetic data generated by different methods}
\end{figure}

\begin{table}[ht!]
\centering
\caption{MAE and RMSE results achieved with different methods on MNIST regression}
\label{tab:table5}
\begin{tabular}{@{}lllll@{}}
\hline
\textbf{Metrics} & \textbf{Teacher Model} & \textbf{Random Sampling} & \textbf{Generator} & \textbf{Direct Optimizer} \\ \hline
MAE              & 0.157                  & \begin{tabular}[c]{@{}l@{}}1.554\\ ± 0.014\end{tabular}  & \begin{tabular}[c]{@{}l@{}}2.422\\ ± 0.027\end{tabular}  & \textbf{\begin{tabular}[c]{@{}l@{}}1.179\\ ±   0.132\end{tabular}} \\
RMSE             & 0.165                  & \begin{tabular}[c]{@{}l@{}}2.872\\ ± 0.063\end{tabular}  & \begin{tabular}[c]{@{}l@{}}2.880\\ ± 0.028\end{tabular}  & \textbf{\begin{tabular}[c]{@{}l@{}}1.978\\ ±   0.143  \end{tabular}}   \\ \hline                 
\end{tabular}
\end{table}
We can observe that the generator method performs only slightly better than random prediction. The direct optimization method was able to provide a substantial improvement in performance compared to the other methods. We examine samples of the synthetic data generated by each method in Figure \ref{fig:fig9}. It is observed that the direct optimization method generates synthetic data closer to what appears to be handwritten digits compared to the other methods. We also examine the histograms of predicted values by teacher networks on a batch of 50 synthetically generated data in Figure \ref{fig:fig10}. It is observed that direct optimization method generates samples with the most diversity while maintaining closeness to integer values. The closer resemblance to real data distribution is likely the reason the student model trained on those synthetic data distills more useful knowledge from the teacher model and outperforms the other methods.

This experiment demonstrates that the direct optimization method can be easily and effectively adapted for data-free knowledge distillation for regression tasks on image inputs, different types of student or generator loss functions and for multilayer networks with non-MLP architectures which potentially addresses the limitation raised in \cite{Kang2021} on poor applicability of the generator method for data-free knowledge distillation of multilayer networks for regression.

\subsection{Case study of data-free distillation for protein solubility predictions}

We experimented with different settings of $\epsilon$ value and found that a value of 0.05 encourages synthetic data generated to be diverse and provides the best training results.
\par
Table \ref{tab:table6} shows the comparison of root mean squared error (RMSE) for the teacher model. The performance obtained for the teacher model is comparable with those obtained in the original study \cite{Han2019} on regression predictive model for protein solubility. Results are averaged over 5 runs. Note that for this dataset, a random model that outputs values uniformly drawn from 0 – 1 will give a RMSE of approximately 0.43.

\begin{table}[ht!]
\centering
\caption{RMSE results achieved with different methods on protein solubility prediction}
\label{tab:table6}
\begin{tabular}{lll}
\hline
\textbf{Teacher Model (SVM)} & \textbf{Random Sampling}                                & \textbf{Direct optimizer}                                        \\ \hline
0.250                        & \begin{tabular}[c]{@{}l@{}}0.287\\ ± 0.001\end{tabular} & \textbf{\begin{tabular}[c]{@{}l@{}}0.267\\ ± 0.005\end{tabular}} \\ \hline
\end{tabular}
\end{table}

It is observed that direct optimization method with differential evolution outperforms random sampling significantly and approaches the RMSE of the teacher model. This case study demonstrated that direct optimization can be easily and effectively applied to cases where gradient information from teacher model is not available, or if the teacher model is not a differentiable neural network at all, such as the support vector machine teacher model in this case. This is achieved by simply swapping the gradient descent with a metaheuristics algorithm for direct optimization. Whereas, using a conventional neural network generative model for synthetic data generation is not possible as the training for the generative model relies on gradients of both the teacher and student model.
\par
The limitation however is that metaheuristics optimization methods tend to be much slower than gradient based optimization and thus incur and significant increase in runtime over the baseline method during training. This may be improved by using a faster metaheuristics algorithm, or one that has been optimized to run on GPU, but that is beyond the scope of this paper.

\section{Conclusion}

In this study, we investigated the behavior of various synthetic data generation methods including random sampling and using an adversarially trained generator. We propose a straightforward synthetic data generation strategy that optimizes the difference between the student and teacher model predictions directly, with additional flexibility to incorporate arbitrary regularization terms that capture properties of the data. We show that synthetic data generated by an adversarially trained generator tends not to represent underlying data distribution well, requiring the need to supplement training with random samples and balancing the loss contributions. Our proposed strategy of direct optimization generates synthetic data with higher loss than random samples while deviating less from underlying distribution than the generator method.
\par
The proposed method allows the student model to learn better and emulate the performance of the teacher model more closely. This is demonstrated in the experiments, where the proposed method achieves lower RMSE than baseline and generator method for most regression datasets tested. We also demonstrate the applicability and flexibility of the method applied to image inputs and deeper convolutional networks on the MNIST dataset, as well as performing distillation on a non-differentiable model in the case study for predicting protein solubility. Nevertheless, there could be many other applications of data-free knowledge distillation for regression with different input data types that we have not investigated. Future work can be done to investigate the performance of this and previous methods on data types such as signal and time series data which can be useful for deployment on small mobile or microcontroller devices for industrial monitoring applications \cite{Kumar2022, Lou2022}. We have used fixed step gradient descent and differential evolution for direct optimization of student loss. Future work can also explore how different optimizers and their settings may affect performance.
\par
We hope that this study furthers the understanding of data-free distillation for regression and highlights the key role of the synthetic data generation process in allowing the student model to effectively distill the teacher model. All codes and data used in this study are available at \url{https://github.com/zhoutianxun/data_free_KD_regression}.

\bibliographystyle{apalike}
\bibliography{reference}

\section*{Supplementary Materials} \label{supplementary}
\subsection*{Guarantee (b) for direct optimization method with fixed steps:}

Assuming a $d$-dimensional $K$-Lipschitz continuous loss function, direct optimization of $x$ with gradient descent in fixed number of steps $t$ on the loss function results in $x_t$ that is bounded in distance away from $x$.
\par
Proof:

Given a $d$-dimensional $K$-Lipschitz function, by definition, for all real $x_1$ and $x_2$:
\[
\lvert f(x_1) - f(x_2) \lvert \leq K \lvert x_1 - x_2 \lvert
\]
Let $x_1 = x_2 + \delta x$ and taking the limits of $\delta x \to 0$:
\[
\lim_{\delta x \to 0} \left| \frac{f(x_2 + \delta x) - f(x_2)}{\delta x} \right| \leq K
\]
\[
\lvert \nabla f \lvert \leq K
\]
For gradient descent, the update formula is:
\[
x_{t+1} = x_t - \eta \nabla f(x_t)
\]
The squared distance moved in the first step of gradient descent is bounded:
\[
\|x_1 - x_0 \|^2 = (\eta \|\nabla f(x_0) \|)^2
\]
\[
\|x_1 - x_0 \|^2 \leq \eta^2 d K^2
\]
Therefore, the distance moved in $t$ steps of gradient descent where $t \geq 1$ on a $d$-dimensional $K$-Lipschitz function is bounded by:
\[
\|x_t - x_0 \| \leq \eta t \sqrt{d} K
\]

\subsection*{Guarantee (b) for generator method with L2 regularization:}

Assuming a $d$-dimensional $K$-Lipschitz continuous loss function, generating synthetic data with a generator network trained with the $L_2$ regularized loss function [equation \ref{eq:3}] will result in $x_g$ that is bounded in distance away from 0, i.e. $L_2$ norm of $x_g$.
\par
Proof:
Given a $d$-dimensional $K$-Lipschitz continuous loss function, the gradient at any point is bounded by $K$:
\[
-K \leq \nabla f \leq K
\]

The generator network is trained to map random noise vector $z$ to $x_g$ to minimize the loss function with a $L_2$ regularization on $x_g$, i.e
\[
L_G(z) = \mathbb{E}_{x_g \sim G_{\phi}(z)}[f(x_g) + \beta \|x_g\|^2]
\]
As we are minimizing the loss function, we only focus on the lower bound on $f(x_g)$
\[
-K \leq \nabla f
\]
Then assuming sufficiently small learning rate, the solution $x_g$ will converge upon a point where gradient of the loss function is 0:
\[
\frac{\partial}{\partial x_g}(f(x_g) + \beta \|x_g\|^2) = 0
\]
\[
\frac{\partial}{\partial x_g}f(x_g) = - 2 \beta x_g
\]
\[
-K \leq - 2 \beta x_g
\]
\[
x_g \leq \frac{K}{2 \beta}
\]
Therefore, the bound for the $L_2$ norm of $x_g$ is:
\[
\|x_g\| \leq \frac{\sqrt{d} K}{2 \beta}
\]

\end{document}